\documentclass{article}

\usepackage{PRIMEarxiv}

\usepackage[utf8]{inputenc} 
\usepackage[T1]{fontenc}    
\usepackage{hyperref}       
\usepackage{url}            
\usepackage{booktabs}       
\usepackage{amsfonts}       
\usepackage{nicefrac}       
\usepackage{microtype}      
\usepackage{lipsum}
\usepackage{fancyhdr}       
\usepackage{graphicx}       
\graphicspath{{media/}}     
\usepackage{amsmath}
\usepackage{amssymb}
\usepackage{mathtools}
\usepackage{amsthm}
\usepackage{enumitem}

\title{Self-Distillation Improves DNA Sequence Inference
}

\author{
  Tong Yu, Lei Cheng, Ruslan Khalitov, Erland Brandser Olsson, Zhirong Yang \\
  Norwegian University of Science and Technology \\
  \texttt{\{tong.yu, lei.cheng, ruslan.khalitov, erland.b.olsson, zhirong.yang\}@ntnu.no} 
}

\begin{document}
\maketitle

\begin{abstract}
Self-supervised pretraining (SSP) has been recognized as a method to enhance prediction accuracy in various downstream tasks. However, its efficacy for DNA sequences remains somewhat constrained. This limitation stems primarily from the fact that most existing SSP approaches in genomics focus on masked language modeling of individual sequences, neglecting the crucial aspect of encoding statistics across multiple sequences. To overcome this challenge, we introduce an innovative deep neural network model, which incorporates collaborative learning between a `student' and a `teacher' subnetwork. In this model, the student subnetwork employs masked learning on nucleotides and progressively adapts its parameters to the teacher subnetwork through an exponential moving average approach. Concurrently, both subnetworks engage in contrastive learning, deriving insights from two augmented representations of the input sequences. This self-distillation process enables our model to effectively assimilate both contextual information from individual sequences and distributional data across the sequence population. We validated our approach with preliminary pretraining using the human reference genome, followed by applying it to 20 downstream inference tasks. The empirical results from these experiments demonstrate that our novel method significantly boosts inference performance across the majority of these tasks. Our code is available at \href{}{https://github.com/wiedersehne/FinDNA.}
\end{abstract}


\section{Introduction}

Masked language modeling (MLM) has experienced significant advancements in the field of natural language processing (NLP) in recent years. Numerous large language models, which have substantially enhanced a variety of language tasks, are based on MLM~\cite{ji2021dnabert,liu2019roberta,brown2020language}. Building on this success in NLP, MLM with self-supervised pretraining is gaining increasing recognition in genomics. This is due to the perception of DNA sequences as having language-like properties within their sequence codes~\cite{zhou2023dnabert,cdilDNA23,avsec2021effective}. The application of language model analogs in genomics pretraining has proven advantageous in several downstream applications, including taxonomic classification~\cite{rizzo2015deep,yu2022paramixer,khalitov2022chordmixer}, enhancer prediction~\cite{sethi2020supervised,yang2017biren}, variant effect prediction~\cite{lee2015method,avsec2021effective}, and gene expression prediction~\cite{zhou2015predicting,avsec2021effective,kelley2018sequential,kelley2020cross}.


However, current MLM-based self-supervised pretraining methods primarily focus on learning contextual information from individual sequences, but they lack the ability to leverage information from other sequences~\cite{tian2019contrastive}. To overcome this limitation, we introduce a sel\textbf{F}-dist\textbf{i}llatio\textbf{n}-based SSL method for \textbf{DNA} (FinDNA) sequence modeling. FinDNA enhances MLM-based self-supervised learning by integrating contrastive learning, which extracts distributional information from the sequence population. Our proposed neural network consists of two components: student and teacher subnetworks. These subnetworks process two different augmented views of input sequences and engage in contrastive learning using sequence-wide representations. Concurrently, the student subnetwork undertakes masked nucleotide learning through position-dependent representation, while the teacher network's weights are updated from the student's via an exponential moving average. This self-distillation method enables learning of data representation that encompasses both contextual and population-based information.


We have pretrained the FinDNA model exclusively on the human reference genome and evaluated it on a broad range of downstream tasks. As anticipated, our method demonstrates improvements over existing state-of-the-art techniques in most human-related inference tasks. For instance, FinDNA surpasses the recent pretraining method HyenaDNA by 22.60\% in accuracy for the Human Regulatory category in GenomicBenchmarks. Remarkably, the pretrained model also shows substantial performance enhancements in other organisms, including mouse enhancers and the COVID virus. Overall, we assert that self-distillation can broadly enhance DNA sequence inference.


The remainder of this paper is structured as follows. Section \ref{sec:notations} outlines the notations and related work. Subsequently, we introduce the FinDNA method in Section \ref{sec:Method}, detailing its network architecture, learning objectives, augmentation techniques, pretraining, fine-tuning, and inference procedures. Section \ref{sec:exp} describes the experimental settings and results. Finally, Section \ref{sec:con} concludes the paper and discusses future research directions.

\section{Notations and Related Work}
\label{sec:notations}

A DNA sequence, consisting of nucleotides, is represented by a string of Roman letters, each denoting one of four nitrogen-containing nucleobases: cytosine (C), guanine (G), adenine (A), or thymine (T). We also introduce the letter `N' to signify ``any one base'' or other ambiguous cases in DNA sequencing due to technical limitations.

In machine learning, vector or tensor inputs are often required. A common approach for nucleobases is one-hot encoding. Here, A, C, G, and T are encoded as $[1\ 0\ 0\ 0\ 0]$, $[0\ 1\ 0\ 0\ 0]$, $[0\ 0\ 1\ 0\ 0]$, and $[0\ 0\ 0\ 1\ 0]$ respectively, with N represented by $[0\ 0\ 0\ 0\ 1]$. After encoding, a DNA sequence of length $L$ is transformed into an $L \times D$ matrix, where $D=5$. It's important to note that this encoding is a raw data representation and does not capture the contextual or distributional nuances of DNA sequences.

The aim of self-supervised pretraining is to derive meaningful representations from extensive unlabeled data, thereby reducing reliance on labeled data in inference tasks. This approach is particularly beneficial in genomics, where labels are often limited.

Most current self-supervised methods for DNA sequences are inspired by masked language models. These models, like DNABERT \cite{ji2021dnabert}, Nucleotide Transformer \cite{dalla2023nucleotide}, and DNABERT-2 \cite{zhou2023dnabert}, mask portions of the DNA sequences and predict these using the unmasked segments, often using Transformers as the backbone for signal integration. Other models like cdilDNA ~\cite{cdilDNA23} and HyenaDNA \cite{nguyen2023HyenaDNA} use different backbones. However, these methods mainly focus on individual sequences, overlooking insights that could be obtained from inter-sequence relationships within populations. Additionally, traditional methods often rely on complex encoding like k-mer or byte pair encoding \cite{sennrich2015neural}, necessitating intricate masking during pretraining to prevent data leakage.

Contrastive learning, another self-supervised approach, has been effective in fields like text \cite{aberdam2021sequence}, images \cite{caron2021emerging}, music \cite{spijkervet2021contrastive}, and time series \cite{poppelbaum2022contrastive}. Methods like SimCLR\cite{chen2020simple}, MoCo \cite{chen2020improved}, and SwAV \cite{caron2020unsupervised} generate positive and negative data pairs through augmentation or transformation, then train models to distinguish between these pairs. This technique extracts distributional data but often misses contextual details within individual items and can be computationally expensive due to the large number of negative pairs.

Self-distillation, a newer method, merges the strengths of masked and contrastive learning. Originating from Knowledge Distillation \cite{hinton2015distilling}, it involves periodically updating a teacher neural network from a student network, as opposed to using a fixed teacher model. This method has been successfully applied in image \cite{zhou2021ibot, oquab2023dinov2} and text processing \cite{aberdam2021sequence}. We will explore its application in DNA sequence analysis next.

\begin{figure*}[!t]
	\begin{center}

  \includegraphics[width=0.92\textwidth]{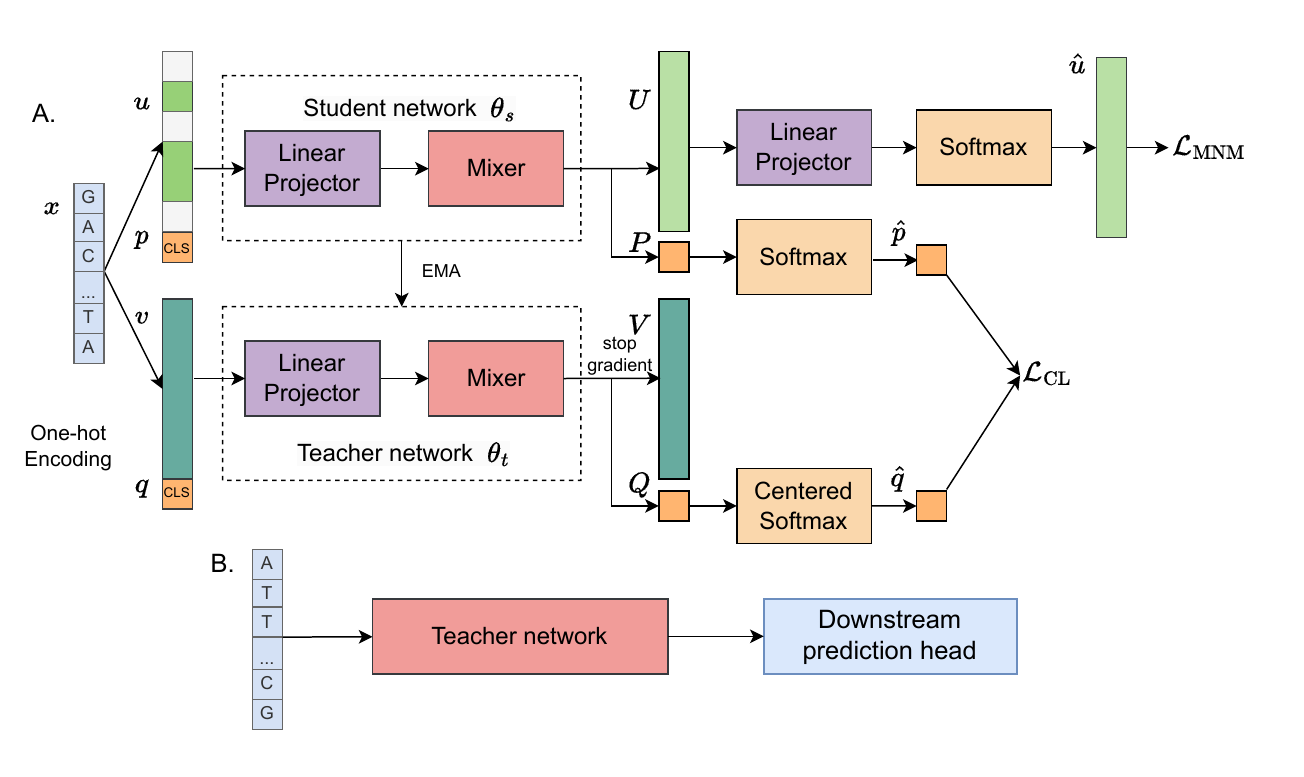}
	\end{center}
 \vspace{-4mm}
	\caption{Illustration of our self-supervised learning model: A. pretraining and B. fine-tuning and inference.}
    \label{fig:FinDNA}
\end{figure*}

\section{Self-Distillation for DNA}
\label{sec:Method}

Conventional self-supervised pretraining to DNA utilizes only information from individual sequences. Here we propose to incorporate self-distillation to extract both contextual information within individual sequences and distributional information among the sequence population.

\subsection{Network architecture}
Our proposed model is shown in Figure \ref{fig:FinDNA}. During pretraining, a DNA sequence is first augmented to two different views $u$ and $v$ (see Section \ref{sec:aug} for augmentation methods). We append $K$ blank [CLS] tokens to the two augmented views (denoted by $p$ and $q$, respectively). After appending, the two sequences are fed to the student and teacher neural networks to obtain their latent representations $[U, P]$ and $[V, Q]$, respectively. The student and teacher have the same network architecture, which includes a linear projector to convert the $D$-dimensional one-hot encoded tokens to $I$-dimensional space and a mixer network that mixes the signals among the tokens.

A conventional choice for the mixer is a stack of Transformers \cite{ji2021dnabert, zhou2023dnabert}. In this work, we adopt ChordMixer \cite{khalitov2022chordmixer}, a recent mixer model that is more scalable to long sequences, in both student and teacher. Each ChordMixer block comprises a simple rotation and an MLP transformation. A ChordMixer layer consists of $\log L$ ChordMixer blocks for a length-$L$ sequence. Every output number can receive information from all input numbers after a ChordMixer layer, which substantially reduces the computational cost compared to a Transformer stack.

\subsection{Learning Objective}
The FinDNA pretraining minimizes the loss $\mathcal{L}$ that combines two pivotal elements, Masked Nucleotide Modeling (MNM) and Contrastive Learning (CL):
\begin{align}
    \mathcal{L} = \alpha\mathcal{L}_{\text{MNM}} + (1-\alpha)\mathcal{L}_{\text{CL}},
\end{align}
where $\alpha\in(0,1)$ controls the tradeoff between the two terms.

The first term, $\mathcal{L}_\text{MNM}$, aims to learn the contextual features within each DNA sequence. We mimic the masked language modeling by treating each nucleotide as tokens in a sequence. A portion of tokens are randomly masked, and the masked sequence is fed to a (student) neural network to infer the masked tokens. Depicted in Figure \ref{fig:FinDNA} Panel A, the upper branch outputs the latent representations $U$ that encode the position-dependent information of an augmented view of the DNA sequence. The tensor $U$ is further processed by a linear projector and a softmax function to obtain a probabilistic prediction of the masked tokens $\hat{u}$. The MNM learning is to minimize the cross-entropy (CE) between $u$ and $\hat{u}$:
\begin{align}
     \mathcal{L}_{\text{MNM}} &= \sum_{j=1}^{M} \text{CE} \left(u^{(j)}|| \hat{u}^{(j)}\right)\\
     &=-\sum_{j=1}^{M}\sum_{d=1}^Du^{(j)}_d\log\hat{u}^{(j)}_d,
\end{align}
where $M$ is the total number of masked nucleotides. Note that we use direct one-hot encoding instead of k-mer or BPE \cite{zhou2023dnabert}, which enables more straightforward masked learning of nucleotides.

The second term, $\mathcal{L}_\text{CL}$, focuses on learning the distributional information within the sequence population. This is implemented by using $P$ and $Q$, the latent representations of the [CLS] tokens that collect sequence-wide information from two different views of the same sequence. Denote $P^{(k)}$ and $Q^{(k)}$ the $k$-th output [CLS] tokens from the student and teacher networks, respectively. We apply softmax to get their probabilistic version:
\begin{align}
    \hat{p}^{(k)}_i &= \frac{\exp\left(P^{(k)}_i/\tau_s\right)}{\sum_{l=1}^K \exp\left(P^{(l)}_i/\tau_s\right)}\\
    \label{eq:teacher_softmax_centering}
    \hat{q}^{(k)}_i &= \frac{\exp\left(\left(Q^{(k)}_i-\xi_{ki}\right)/\tau_t\right)}{\sum_{l=1}^K \exp\left(\left(Q_d^{(l)}-\xi_{li}\right)/\tau_t\right)},
\end{align}
where $K$ is the number of [CLS] tokens, and $\tau_s$ and $\tau_t$ are temperature hyperparameters that control the sharpness of the output distributions ($\tau_t<\tau_s$ in practice). 

In Eq.~\ref{eq:teacher_softmax_centering}, we subtracted the same vector $\xi$ from each teacher network output [CLS] token. The matrix $\xi\in\mathbb{R}^{K\times D}$ is updated as
\begin{align}
\xi\leftarrow \beta \xi + (1-\beta)c,
\end{align}
where $\beta\in(0,1)$ is the forgetting factor and $c$ is the center of all $Q$'s in the same batch. Such moving-averaged centering leads to normalization within the batch and can prevent students and teachers from collapsing into constant networks.
The contrastive learning objective can then be calculated as
\begin{align}
    \mathcal{L}_\text{CL}&=\sum_{k=1}^K \text{CE}\left(\hat{q}^{(k)}||\hat{p}^{(k)}\right)\\
    &=-\sum_{k=1}^K \sum_{i=1}^I\hat{q}^{(k)}_i\log \hat{p}^{(k)}_i
\end{align}

\subsection{DNA Sequence Augmentation}
\label{sec:aug}

Nicholas et al. \cite{lee2023evoaug} proposed a variety of augmentation functions specifically designed for DNA sequences. This work considers the following functions:
\begin{itemize}[topsep=0pt]
\setlength\itemsep{1pt}
    \item \emph{Translocation}: This function involves selecting a random breakpoint in a DNA sequence. The sequence is then split into two segments, which are subsequently swapped.

    \item \emph{Insertion}: This function entails the random insertion of a DNA fragment (whose length varies) at a randomly chosen position within the wild-type sequence.

    \item \emph{Deletion}: In this approach, a randomly chosen contiguous segment is removed from the wild-type sequence. To maintain the original length, the resulting shorter sequence is padded with a random DNA sequence.

    \item \emph{Reverse-Complement}: This process involves substituting the entire sequence with its reverse-complement at a certain probability.

    \item \emph{Gaussian Noise}: Gaussian noise is introduced into the sequence. The distribution of this noise is defined by parameters noise\_mean and noise\_std, which are applied to the input sequence. In this paper, we set noise\_mean as 0 and noise\_std as 0.2.

    \item \emph{Masking}: This augmentation method randomly obscures a specific proportion of nucleotides in the input sequence. We have used a masking ratio of 30\% and represented the masks as $[0\ 0\ 0\ 0\ 0]$.
\end{itemize}


\subsection{Pretraining, Fine-tuning and Inference}
The pretraining procedure is shown in Figure \ref{fig:FinDNA} Panel A. 
In self-distillation, only the student network is learned by gradient back-propagation. The teacher network is updated by an exponential moving average (EMA) from the student:
\begin{align}
    \theta_t \leftarrow \lambda\theta_t + (1-\lambda)\theta_s
\end{align}
where $\theta_s$ and $\theta_t$ are the neural network weights of the student and teacher, respectively, and $\lambda\in(0,1]$ is the forgetting factor. In this way, the learned information from the student network can be distilled into the teacher network. 

Panel B of Figure \ref{fig:FinDNA} illustrates the fine-tuning and inference stages of FinDNA. The input sequence without augmentation is fed to the pretrained teacher network. The resulting latent representation ($V$ and/or $Q$) is used for a downstream prediction task. The downstream prediction head is a lightweight neural network, e.g., a linear layer or a two-layer MLP, for classification, regression, etc. In fine-tuning, both the teacher network and the prediction head are updated using gradient backpropagation.

\section{Experiments}
\label{sec:exp}

To verify that self-distillation improves DNA sequence inference, we have pretrained our model with the human reference genome and performed extensive tests on in total 20 DNA sequence-based inference tasks from three different benchmark sources: 1) GenomicBenchmarks \cite{grevsova2023genomic}, 2) Genome Understanding Evaluation \cite{zhang2023dnagpt}
, and 3) MTcDNA \cite{yu2022paramixer}.
The experiments were conducted on a Linux machine with 8$\times$NVIDIA Telsa A100-40GB GPUs.



\subsection{Pretraining}

We collected 100,000 DNA sequences with a sequence length of 1000bp from human reference genome (GRCh38) for pretraining. Two different augmentations were applied on an input sequence to get $u$ and $v$, where we used a combination of random deletion, random insertion, random translocation, and random masking for $u$, and Gaussian noise followed by reverse-complement for $v$. 

After augmentation, ten [CLS] tokens are appended to each DNA sequence (i.e., $K=10$). To train the student network, a cosine scheduler is employed for the learning rate, with 30\% of training steps allocated for warming up. We trained the model for 50 epochs in total. 

In self-distillation, the student and teacher subnetworks operate at temperatures of 0.1 and 0.04, respectively. The updates of $\lambda$ follow a cosine scheduler, ranging from 0.996 to 1. The forgetting factor for center updates $\beta$ was set at 0.996. We set the tradeoff $\alpha=0.5$ between $\mathcal{L}_\text{MNM}$ and $\mathcal{L}_\text{CL}$. For the mixer network, we have used four ChordMixer layers, each with 308 channels and 512 hidden dimensions.

\begin{table*}[t]
\small
\begin{center}
\begin{tabular}{lcccc}
\hline\hline\\[-3.5mm]
Dataset & Mouse Enhancers & Coding vs Intergenomic & Human vs Worm &  Human Enhancers Cohn\\
\hline\\[-3mm]
HyenaDNA & $76.86 \pm 0.31$ & $86.23 \pm 0.01$ & $82.63 \pm 0.02$ & $68.23 \pm 0.12$  \\ [1mm] FinDNA-T & $78.23 \pm 0.05 $ & $85.80 \pm 0.01 $ & $74.35 \pm 0.15 $ & $67.73 \pm 0.17 $  \\ [1mm]
FinDNA & $\textbf{78.43} \pm \textbf{0.22}$ & $\textbf{87.79} \pm \textbf{0.02}$ & $\textbf{91.30} \pm \textbf{0.15}$ & $\textbf{70.04} \pm \textbf{0.05}$ \\ [1mm]
\hline\\[-3mm]
\end{tabular}

\begin{tabular}{lcccc}
\hline\hline\\[-3.5mm]
Dataset  & Human Enhancers Ensembl & Human Regulatory & Human Nontata Promoters & Human OCR Ensembl\\
\hline\\[-3mm]
HyenaDNA & $66.18 \pm 0.55$ & $57.42 \pm 0.01$ & $80.01 \pm 0.08$ & $62.52 \pm 0.03$  \\ [1mm] 
FinDNA-T & $66.17 \pm 0.24$ & $80.22 \pm 0.05$ & $82.89 \pm 0.05$ & $66.71 \pm 0.11$\\ [1mm]
FinDNA & $\textbf{67.90} \pm \textbf{0.03}$ & $\textbf{80.02} \pm \textbf{0.34}$ & $\textbf{83.55} \pm \textbf{0.17}$ & $\textbf{66.90} \pm \textbf{0.04}$ \\ [1mm] 
\hline\\[-3mm]
\end{tabular}
\end{center}
\caption{Top-1 accuracy ($\times100\%$) for GenomicBenchmarks using linear probing (with pretrained models frozen). Boldface numbers indicate the best. FinDNA-T denotes FinDNA framework with Transformer as the backbone network. The mean and standard deviation are calculated based on five runs of each task.}
\label{tab:genomic}
\vspace{2mm}
\end{table*}

\subsection{Compared methods}
We compare FinDNA with four recent models for DNA sequence-based inferences: DNABERT \cite{ji2021dnabert}\footnote{https://github.com/jerryji1993/DNABERT}, DNABERT-2 \cite{zhou2023dnabert}\footnote{https://github.com/Zhihan1996/DNABERT\_2}, Nucleotide Transformer\cite{dalla2023nucleotide} \footnote{https://github.com/instadeepai/nucleotide-transformer}, and HyenaDNA \cite{nguyen2023HyenaDNA}\footnote{https://github.com/HazyResearch/hyena-dna}. All hyperparameters of the compared methods were tuned by using cross-validation.

\textbf{DNABERT} mimics the BERT language model based on Transformers by using $k$-mers as sequence tokens. There are four versions of DNABERT ($k=3,4,5,6$). Here we compared our method with the best reported DNABERT ($k=6$).

\textbf{DNABERT-2} was a more recent release of DNABERT. It was trained on human genome and the multi-species genome. The method applies BPE and FlashAttention \cite{dao2022flashattention} to improve the model capacity and efficiency.

In the comparison, we also include its variant DNABERT-2 \(\blacklozenge\), which uses extra pretraining on downstream task data and reports better results than DNABERT-2 on some tasks.

\textbf{HyenaDNA} was trained on human genome sequences at single nucleotide resolution. The model uses a decoder-only architecture defined by a stack of blocks consisting of a Hyena operator \cite{poli2023hyena} for long-range convolutions. We have used the same input length (1k) in HyenaDNA as in FinDNA and other compared methods.

\textbf{Nucleotide Transformer} is a collection of models pretrained on DNA sequences. They employ Transformer as the backbone and were pretrained using 3,202 genomes from human and 850 genomes from other species. The models have different numbers of parameters, ranging from 50M to 2.5B. We compare our model with NT-500m-human and NT-2500m-multi, the two with the best performance reported in their paper.

\begin{table*}[t]
\begin{center}
\begin{tabular}{lccccc}
\hline\hline\\[-3.5mm]
Dataset     &   DNABERT  &  HyenaDNA      & DNABERT-2    &  CM-MNM      & FinDNA \\
\hline\\[-3mm]
Mouse Enhancers & $76.86 \pm 1.17$ & $84.25 \pm 0.05$  & \underline{$84.85 \pm 0.38$}   & $83.42 \pm 0.02$ & $\textbf{85.55} \pm \textbf{0.03}$\\ [1mm]

Coding vs Intergenomic & $88.24 \pm 0.20$  & $88.09 \pm 0.03$ & $\textbf{94.19} \pm \textbf{0.03}$ & $93.53 \pm 0.05$  & $\underline{93.73 \pm 0.01}$ \\ [1mm]
 
Human vs Worm & $95.60 \pm 0.01$ & $96.70 \pm 0.18$ & $\textbf{97.74} \pm \textbf{0.04}$ & $96.33 \pm 0.02$ & $\underline{96.88 \pm 0.12}$\\ [1mm]

Human Enhancers Cohn & $65.96 \pm 1.15$  & $\underline{73.75} \pm 0.02$ & $73.13 \pm 0.17$ & $73.30 \pm 0.09$ & $\textbf{74.20} \pm \textbf{0.03}$ \\ [1mm]
Human Enhancers Ensembl & $83.97 \pm 0.08$ & $89.41 \pm 0.09$ & \underline{$92.77 \pm 0.11$} & $92.15 \pm 0.08$ & $\textbf{93.30} \pm \textbf{0.01}$\\ [1mm]
Human Regulatory & $91.47 \pm 1.58$ & $93.80 \pm 0.01$ & $91.73 \pm 0.90$ & $\underline{93.85 \pm 0.03}$ & $\textbf{93.88} \pm \textbf{0.02}$ \\ [1mm]
Human Nontata Promoters & $94.52 \pm 1.01$ & $96.65 \pm 0.03$ & $95.00 \pm 0.37$ & $\underline{97.13 \pm 0.17}$  & $\textbf{97.39} \pm \textbf{0.04}$\\ [1mm]
Human OCR Ensembl & $78.55 \pm 0.30$ & $\underline{80.29} \pm 0.19$ & $79.49 \pm 0.40$  & $78.23 \pm 0.25$& $\textbf{81.19} \pm \textbf{0.12}$\\ [1mm]
\hline\\[-3mm]
\hline
\end{tabular}
\end{center}
\caption{
Top-1 accuracy ($\times100\%$) of various models on GenomicBenchmarks (with pretrained models finetuned). Boldface numbers indicate the best, and the underlined numbers refer to the runner-ups. CM-MNM is a variant of our method for ablation study, where it replaces self-distillation with conventional masked learning over nucleotides (i.e., using ChordMixer backbone and only the $\mathcal{L}_\text{MNM}$ loss). Means and standard deviations were derived from the results of five independent runs.}
\label{tab:genomic-ft}
\end{table*}

\begin{table*}[t]
\small
\begin{center}
\begin{tabular}{lcccccc}
\hline\hline\\[-3.5mm]
\multicolumn{1}{c}{} & \multicolumn{6}{c}{\textbf{Epigenetic Marks Prediction}}\\
\hline\\[-3mm]
Dataset & H3 & H3K14ac & H3K36me3 &  H3K4me1 & H3K4me2 & H4\\
\hline\\[-3mm]
DNABERT (6-mer) & 74.15 & 40.06 & 47.25 & 41.44  & 32.27 & 79.26\\ [1mm]
NT-500M-human & 69.67 & 33.55 & 44.14 & 37.15 & 30.87 & 76.17\\ [1mm]
NT-2500M-multi & \underline{78.77} & 56.20 & \underline{61,99} & \underline{55.30} & 36.49 & \underline{81.67}\\ [1mm]
DNABERT-2 & 78.27 & 52.57 & 56.88 & 50.52 & 31.13 & 80.71\\ [1mm]
DNABERT-2 \(\blacklozenge\) & \textbf{80.17} & \underline{57.42} & 61.90 & 53.00 & \underline{39.89} & \textbf{81.86} \\ [1mm]
\hline\\[-3mm]
FinDNA  & $77.81 \pm 0.38$ & $\textbf{66.41} \pm \textbf{0.05}$ & $ \textbf{66.69} \pm \textbf{0.21}$& $\textbf{55.08} \pm \textbf{0.03}$ & $\textbf{50.73} \pm \textbf{0.04}$ & $ 78.93 \pm 0.05$ \\ [1mm]
\hline\\[-3mm]
\end{tabular}

\begin{tabular}{lccccccc}
\hline\hline\\[-3.5mm]
\multicolumn{1}{c}{} & \multicolumn{4}{c}{\textbf{Epigenetic Marks Prediction}} & \multicolumn{1}{c}{\textbf{Virus}} &  \multicolumn{1}{c}{\textbf{Average}} & \multicolumn{1}{c}{\textbf{Param.}} \\
\hline\\[-3mm]
Dataset  & H3K79me3 & H3K9ac & H3K4me3 & H4ac & Covid\\
\hline\\[-3mm]
DNABERT (6-mer) & 61.17 & 51.22 & 27.81 & 37.43 & 62.23 & 50.39 & 89M \\ [1mm]
NT-500M-human & 58.35 & 45.81 & 24.06 & 33.74 & 55.50 & 46.27 & 480M\\ [1mm]
NT-2500M-multi & 64.70 & 56.01 & 40.34 & 49.13 & \underline{73.04} & 59.42 & 2537M \\ [1mm]
DNABERT-2 & \underline{67.39} & 55.63 & 36.27 & 50.43 & 71.02 & 57.34 & 117M\\ [1mm]
DNABERT-2 \(\blacklozenge\) & 65.46 & \underline{57.07} & \underline{41.20} & \underline{50.35} & 68.49 & \underline{59.71} & 117M\\ [1mm]
\hline\\[-3mm]
FinDNA  & $\textbf{72.42} \pm \textbf{0.06}$& $\textbf{64.72} \pm \textbf{0.17}$ & $\textbf{59.80} \pm \textbf{0.15}$& $\textbf{64.73} \pm \textbf{0.26} $& $\textbf{74.09} \pm \textbf{0.09}$& \textbf{66.49} & \textbf{25.4M}\\ [1mm]
\hline\\[-5mm]
\end{tabular}
\end{center}
\caption{Performance comparison for GUE benchmark. We quote the experimental results from DNABERT-2 and report Matthews Correlation Coefficients for Epigenetic Marks Prediction and F1-scores for Virus. Boldface numbers indicate the best results, and the underlined numbers refer to the runner-ups. Note that DNABERT-2 \(\blacklozenge\) used further masked language modeling pretraining on the training sets of every GUE task. Means and standard deviations of FinDNA were derived from the results of five independent runs.}
\label{tab:gue}
\vspace{-1mm}
\end{table*}

\subsection{GenomicBenchmarks}
GenomicBenchmarks is a recently introduced public benchmark, which encompasses eight distinct regulatory element classification tasks. Seven of these tasks involve binary classification, while one task, specifically the Human Regulatory task, involves ternary classification. The sequence lengths within these datasets vary from 2 to 4,776. The median length in each task ranges from 200 to 2,381 (see Appendix Table \ref{tab:genomicdata}). To train our model, we utilized a batch size of 1024, a learning rate of 0.01, and a weight decay of 0.1. The model underwent trained for 50 epochs with 30\% training steps for warm-up.

We first used linear probing to compare the features extracted by pretrained FinDNA, FinDNA-T (a FinDNA framework using Transformer as backbone) and HyenaDNA (a DNA pretraining approach based solely on MLM). In this process, with the pretrained models fixed, the extracted features were input into a linear classifier, which was then trained with the data from each task in GenomicBenchmarks. We adhered to the benchmark's recommended metrics and reported Top-1 Accuracy for all evaluated methods.

The outcomes are presented in Table \ref{tab:genomic}, clearly demonstrating that FinDNA surpasses both FinDNA-T and HyenaDNA across all evaluated datasets. Significantly, FinDNA achieves a performance advantage of 22.81\% and 8.7\% over HyenaDNA in the Human Regulatory and Human vs. Worm tasks, respectively. These results further reveal that the combination of FinDNA with ChordMixer yields superior performance when compared to the Transformer model. Consequently, we have decided to employ ChordMixer in our subsequent experiments.

Furthermore, we fine-tuned FinDNA on the same tasks and conducted comparisons with several MLM-based models. We also assessed FinDNA against CM-MNM---a model using the ChordMixer backbone with an exclusive focus on MNM loss---to determine the impact of self-distillation on DNA sequence modeling. The findings, presented in Table \ref{tab:genomic-ft}, show that our model not only surpasses the state-of-the-art model, HyenaDNA, in performance on six out of eight datasets but also secures the second place on the remaining two. Compared to the extensively pretrained model DNABERT-2, FinDNA achieves superior average accuracy, indicating its robust capability in DNA sequence modeling despite its smaller size. Additionally, FinDNA outperforms CM-MNM in seven out of eight datasets, underscoring the efficacy of the self-distillation strategy in enhancing the performance of models based solely on masked learning.

\subsection{GUE Benchmarks}

The Genomic Underpinning Evaluation (GUE) includes seven extensive genome sequence classification challenges, spread over 28 datasets, designed for tasks involving human, mouse, virus, and yeast species. Our research primarily targets two demanding tasks within GUE: Epigenetic Marks Prediction (EMP) and Virus classification with extended sequence lengths. EMP comprises 10 datasets, each aiming to predict the presence of epigenetic marks within a 500bp DNA sequence. In contrast, the Virus classification task focuses on the Covid dataset, which requires identifying nine distinct Covid variants from virus DNA sequences measuring 1kbp in length. Our approach adheres to the benchmark set in DNABERT-2, employing Matthew's Correlation Coefficient (MCC) for EMP and the F1-Score for Virus classification.

We used cross-validation to determine the hyperparameters for both the EMP and Virus tasks. For EMP, the model configuration included a batch size of 32 and a learning rate of $5\times10^{-4}$, with the model being fine-tuned over 20 epochs to achieve optimal validation performance. For Virus classification, we chose a batch size of 256 and a learning rate of 0.001, training the model across 100 epochs. Both tasks utilized a cosine scheduler for learning rate adjustment, with a 30\% warm-up phase. Additionally, we maintained a dropout rate of 0.1 for both the MLP and rotation layers in the ChordMixer model.

The results, as presented in Table \ref{tab:gue}, highlight FinDNA's exceptional performance. FinDNA outshines in eight out of ten EMP datasets and achieves the highest scores in the Virus classification task. FinDNA surpasses the runner-up, (DNABERT-2 $\blacklozenge$), by 6.74\% on average. Significant performance boosts include improvements in MCC by 18.73\%, 14.64\%, and 10.8\% for the H3K4me3, H4ac, and H3K4me2 markers, respectively. Remarkably, despite being exclusively pretrained on the human genome, FinDNA outperforms DNABERT-2 $\blacklozenge$ whose pretraining includes the downstream data.

Furthermore, FinDNA showcases enhanced parameter efficiency, requiring significantly fewer parameters compared to other models (22\% of DNABERT-2 and 5\% of NT-500M-human) to deliver superior prediction outcomes. This reduced model size facilitates easier deployment on more cost-effective devices, underscoring FinDNA's practical advantages.

\begin{table*}[t]
\begin{center}
\begin{tabular}{lcccc}
\hline\hline\\[-3.5mm]
Dataset     &   M+NoAug  &  M+DIT      & MDT+DIT    & MDT+NR  \\
\hline\\[-3mm]
KL-divergence & 0.025 & 0.149 & 0.152 & 0.175 \\
\hline\\[-3mm]
Mouse Enhancers & \textbf{80.58}  & 78.1   & 79.75 & 78.51\\ [1mm]

Coding vs Intergenomic &  87.60 & 87.74 & 86.88  & \textbf{87.81}\\ [1mm]
 
Human vs Worm & 87.43 & 90.44 & 89.61 & \textbf{91.34}\\ [1mm]

Human Enhancers Cohn & 69.11 & 69.73 & 69.52 & \textbf{70.09} \\ [1mm]
Human Enhancers Ensembl & 63.76 & 66.64 & \textbf{69.75} & 67.92\\ [1mm]
Human Regulatory & 74.08 & 71.00 & 79.84 & \textbf{80.24} \\ [1mm]
Human Nontata Promoters & 83.31 & 83.17 & 83.37  & \textbf{83.64}\\ [1mm]
Human OCR Ensembl & 66.90 & 67.24  & \textbf{67.41} & 66.98\\ [1mm]
Average & 76.59 & 76.76 & 78.26 & \textbf{78.31} \\ [1mm]
\hline
\end{tabular}
\end{center}
\caption{Ablation Study Results: Evaluating the Impact of Various Augmentation Strategies on GenomicBenchmarks with Linear Probing (Top-1 Accuracy in Percentage). This table explores the efficacy of different combinations of augmentations: M (Random Masking), D (Random Deletion), I (Random Insertion), T (Random Translocation), N (Random Noise), and R (Reverse-completion). The symbols before and after the "+" sign denote the augmentations applied to generate the two augmented views uu and vv, which are then used as inputs for the student and teacher networks, respectively. Boldface numbers indicate the best combination for a task.}
\label{tab:genomic-lp}
\vspace{2mm}
\end{table*}

\setlength{\tabcolsep}{6pt}
\begin{table}[t]
\begin{center}
\begin{tabular}{lccc}
\hline\hline\\[-3.5mm]
Lengths & 1024 & 4096 & 8192\\
\hline\\[-3mm]
NT-500m-human & 89.92 & 92.79 & 93.81 \\ [1mm]
DNABERT-2 & 92.36 & 95.44 & 97.52  \\ [1mm]
\hline\\[-3mm]
FinDNA & \textbf{95.26} & \textbf{98.04} & \textbf{98.82}  \\ [1mm]
\hline
\end{tabular}
\end{center}
\caption{Top-1 accuracy ($\times 100\%$) of MTcDNA for three different lengths (bp).}
\label{tab:MTcDNA}
\end{table}

\begin{table}[t]
\begin{center}
\begin{tabular}{lccc}
\hline\hline\\[-3.5mm]
 & Time & Memory & FLOPs\\
\hline\\[-3mm]
NT-500m-human & $\times 2.98$ & $\times 3.60$ & $\times 8.35$ \\ [1mm]
DNABERT-2 & $\times1.71$ & $\times 2.21$ & $\times 2.75$ \\ [1mm]
\hline\\[-3mm]
FinDNA & $\times1$ & $\times1$ & $\times1$  \\ [1mm]
\hline\\[-3mm]
\end{tabular}
\end{center}
\caption{Factors of consumed time, memory, and FLOPs on the task MTcDNA-1024bp during training.}
\label{tab:flops}
\end{table}

\subsection{MTcDNA Benchmark}

We employed an additional benchmark to evaluate the transfer performance of our proposed model using the MTcDNA dataset, which contains cDNA sequences from mice and turtles and was introduced by Paramixer \cite{yu2022paramixer}. The objective is to classify each cDNA sequence as either mouse or turtle. The mouse category contains 12,300 cDNA sequences from two species, Mus musculus and Mus spretus, while the turtle category consists of 4,193 sequences from Chelonoidis abingdonii and Gopherus agassizii.

For this evaluation, we continued to use the pretrained model based on the human reference genome, which had been pretrained with a sequence length of 1000 base pairs (bp). Furthermore, we deliberately established the MTcDNA classification task at three different sequence lengths: 1024bp, 4096bp, and 8192bp. This approach allowed us to assess 1) the model's capability to integrate distant information effectively for more precise predictions and 2) the impact of pretraining on shorter sequences on the model's performance during fine-tuning on longer sequences.

We compared FinDNA with two other models, DNABERT-2 and NT-500m-human, and reported the top-1 accuracies in Table \ref{tab:MTcDNA}. The results indicate that all models demonstrate enhanced performance as the sequence length increases from 1024bp to 8192bp. Notably, FinDNA consistently outperformed the other models, achieving the highest accuracy at all three tested sequence lengths.

\subsection{Computing Cost Analysis}

The comparison of FinDNA with two alternative models, DNABERT-2 and NT-500m-human, was conducted focusing on the metrics of time consumption, memory usage, and FLOPS during training for the MTcDNA classification task, which involves sequences of 1024 base pairs. Table \ref{tab:flops} presents these consumption factors in relation to FinDNA.  The results demonstrate that FinDNA is the most efficient model in terms of time, memory, and floating-point operations compared to the other two models. We also conducted aplation study on the overhead of using self-distillation technique (See Appendix Table \ref{tab:scales}).

\subsection{Ablation Study on Augmentations}

The FinDNA architecture incorporates two distinct augmented views of a sequence, prompting an investigation into whether increased dissimilarity between these views enhances representation quality.

This dissimilarity is quantified using the Kullback-Leibler (KL) divergence. For two input views $u\in\mathbb{R}^{L\times D}$ and $v\in\mathbb{R}^{L\times D}$, we first normalize each row using a softmax function to produce their probabilistic counterparts $\check{u}$ and $\check{v}$. The dissimilarity between these probabilistic views is then calculated as follows:
\begin{align}
    D_\text{KL}(\check{u}||\check{v})=\sum_{j=1}^L\sum_{d=1}^D\check{u}^{(j)}_d\log\frac{\check{u}^{(j)}_d}{\check{v}^{(j)}_d}
\end{align}
The overall dissimilarity is the sum of the KL divergence across all sequences.

Our research specifically examines the impact of four augmentation pairings, each offering a progressively higher level of dissimilarity: M+NoAug, M+DIT, MDT+DIT, and MDIT+NR, where M stands for Random Masking, D for Random Deletion, I for Random Insertion, T for Random Translocation, N for Random Noise, and R for Reverse-completion. NoAug means no augmentation.

Our experiments were carried out using the GenomicBenchmarks dataset for the above pairs of augmentations. The results, shown in Table \ref{tab:genomic-lp}, reveal a significant difference in dissimilarity values, with the pair MDT and NR showing a much higher dissimilarity (0.175) compared to the M and NoAug pair (0.025). Furthermore, the combination of MDT+NR achieved the highest average accuracy, whereas M+NoAug exhibited the lowest performance across all pairs. These findings support our initial hypothesis, suggesting that augmentations with greater dissimilarity between the two views enhance the effectiveness of downstream inferences. 

\section{Conclusion}
\label{sec:con}

We have proposed an innovative deep neural network model that enhances DNA sequence-based inference through self-supervised pretraining. Our model incorporates self-distillation to learn both contextual information within individual sequences and distributional information across multiple sequences. We have pretrained a neural network using the human reference genome and tested it on 20 various downstream inference tasks sourced from three public benchmarks. Our experimental findings demonstrate that our approach significantly surpasses existing methods in performance.

Looking ahead, our methodology opens the door to more extensive pretraining possibilities. Beyond the human reference genome, we plan to incorporate additional DNA sequences from various organisms. This expansion will include pretraining and inference processes for longer and variable-length DNA sequences. The resulting pretrained model holds immense potential for broader applications such as personalized medicine, genetic mutation analysis, agricultural development, disease pathogen tracking, and epidemiological studies.

\section{Impact Statement}
This paper presents work whose goal is to advance the field of DNA sequence modeling. There are many potential societal consequences of our work, none which we feel must be specifically highlighted here.

\bibliographystyle{unsrt}  
\bibliography{main}  

\begin{thebibliography}{10}

\bibitem{ji2021dnabert}
Yanrong Ji, Zhihan Zhou, Han Liu, and Ramana~V Davuluri.
\newblock {DNABERT}: pre-trained bidirectional encoder representations from {Transformers} model for {DNA}-language in genome.
\newblock {\em Bioinformatics}, 37(15):2112--2120, 2021.

\bibitem{liu2019roberta}
Yinhan Liu, Myle Ott, Naman Goyal, Jingfei Du, Mandar Joshi, Danqi Chen, Omer Levy, Mike Lewis, Luke Zettlemoyer, and Veselin Stoyanov.
\newblock Roberta: A robustly optimized {BERT} pretraining approach.
\newblock {\em arXiv preprint arXiv:1907.11692}, 2019.

\bibitem{brown2020language}
Tom Brown, Benjamin Mann, Nick Ryder, Melanie Subbiah, Jared~D Kaplan, Prafulla Dhariwal, Arvind Neelakantan, Pranav Shyam, Girish Sastry, Amanda Askell, et~al.
\newblock Language models are few-shot learners.
\newblock {\em Advances in neural information processing systems}, 33:1877--1901, 2020.

\bibitem{zhou2023dnabert}
Zhihan Zhou, Yanrong Ji, Weijian Li, Pratik Dutta, Ramana Davuluri, and Han Liu.
\newblock {DNABERT}-2: Efficient foundation model and benchmark for multi-species genome.
\newblock {\em arXiv preprint arXiv:2306.15006}, 2023.

\bibitem{cdilDNA23}
Lei Cheng, Tong Yu, Ruslan Khalitov, and Zhirong Yang.
\newblock Self-supervised learning for dna sequences with circular dilated convolutional networks.
\newblock {\em Neural Networks}, 171:466--473, 2024.

\bibitem{avsec2021effective}
{\v{Z}}iga Avsec, Vikram Agarwal, Daniel Visentin, Joseph~R Ledsam, Agnieszka Grabska-Barwinska, Kyle~R Taylor, Yannis Assael, John Jumper, Pushmeet Kohli, and David~R Kelley.
\newblock Effective gene expression prediction from sequence by integrating long-range interactions.
\newblock {\em Nature methods}, 18(10):1196--1203, 2021.

\bibitem{rizzo2015deep}
Riccardo Rizzo, Antonino Fiannaca, Massimo La~Rosa, and Alfonso Urso.
\newblock A deep learning approach to {DNA} sequence classification.
\newblock In {\em International Meeting on Computational Intelligence Methods for Bioinformatics and Biostatistics}, pages 129--140. Springer, 2015.

\bibitem{yu2022paramixer}
Tong Yu, Ruslan Khalitov, Lei Cheng, and Zhirong Yang.
\newblock {Paramixer}: Parameterizing mixing links in sparse factors works better than dot-product self-attention.
\newblock In {\em Proceedings of the IEEE/CVF Conference on Computer Vision and Pattern Recognition}, pages 691--700, 2022.

\bibitem{khalitov2022chordmixer}
Ruslan Khalitov, Tong Yu, Lei Cheng, and Zhirong Yang.
\newblock {ChordMixer}: A scalable neural attention model for sequences with different lengths.
\newblock {\em arXiv preprint arXiv:2206.05852}, 2022.

\bibitem{sethi2020supervised}
Anurag Sethi, Mengting Gu, Emrah Gumusgoz, Landon Chan, Koon-Kiu Yan, Joel Rozowsky, Iros Barozzi, Veena Afzal, Jennifer~A Akiyama, Ingrid Plajzer-Frick, et~al.
\newblock Supervised enhancer prediction with epigenetic pattern recognition and targeted validation.
\newblock {\em Nature methods}, 17(8):807--814, 2020.

\bibitem{yang2017biren}
Bite Yang, Feng Liu, Chao Ren, Zhangyi Ouyang, Ziwei Xie, Xiaochen Bo, and Wenjie Shu.
\newblock {BiRen}: predicting enhancers with a deep-learning-based model using the {DNA} sequence alone.
\newblock {\em Bioinformatics}, 33(13):1930--1936, 2017.

\bibitem{lee2015method}
Dongwon Lee, David~U Gorkin, Maggie Baker, Benjamin~J Strober, Alessandro~L Asoni, Andrew~S McCallion, and Michael~A Beer.
\newblock A method to predict the impact of regulatory variants from {DNA} sequence.
\newblock {\em Nature genetics}, 47(8):955--961, 2015.

\bibitem{zhou2015predicting}
Jian Zhou and Olga~G Troyanskaya.
\newblock Predicting effects of noncoding variants with deep learning--based sequence model.
\newblock {\em Nature methods}, 12(10):931--934, 2015.

\bibitem{kelley2018sequential}
David~R Kelley, Yakir~A Reshef, Maxwell Bileschi, David Belanger, Cory~Y McLean, and Jasper Snoek.
\newblock Sequential regulatory activity prediction across chromosomes with convolutional neural networks.
\newblock {\em Genome research}, 28(5):739--750, 2018.

\bibitem{kelley2020cross}
David~R Kelley.
\newblock Cross-species regulatory sequence activity prediction.
\newblock {\em PLoS computational biology}, 16(7):e1008050, 2020.

\bibitem{tian2019contrastive}
Yonglong Tian, Dilip Krishnan, and Phillip Isola.
\newblock Contrastive representation distillation.
\newblock {\em arXiv preprint arXiv:1910.10699}, 2019.

\bibitem{dalla2023nucleotide}
Hugo Dalla-Torre, Liam Gonzalez, Javier Mendoza-Revilla, Nicolas~Lopez Carranza, Adam~Henryk Grzywaczewski, Francesco Oteri, Christian Dallago, Evan Trop, Bernardo~P de~Almeida, Hassan Sirelkhatim, et~al.
\newblock The nucleotide transformer: Building and evaluating robust foundation models for human genomics.
\newblock {\em bioRxiv}, pages 2023--01, 2023.

\bibitem{nguyen2023HyenaDNA}
Eric Nguyen, Michael Poli, Marjan Faizi, Armin Thomas, Callum Birch-Sykes, Michael Wornow, Aman Patel, Clayton Rabideau, Stefano Massaroli, Yoshua Bengio, et~al.
\newblock Hyenadna: Long-range genomic sequence modeling at single nucleotide resolution.
\newblock {\em arXiv preprint arXiv:2306.15794}, 2023.

\bibitem{sennrich2015neural}
Rico Sennrich, Barry Haddow, and Alexandra Birch.
\newblock Neural machine translation of rare words with subword units.
\newblock {\em arXiv preprint arXiv:1508.07909}, 2015.

\bibitem{aberdam2021sequence}
Aviad Aberdam, Ron Litman, Shahar Tsiper, Oron Anschel, Ron Slossberg, Shai Mazor, R~Manmatha, and Pietro Perona.
\newblock Sequence-to-sequence contrastive learning for text recognition.
\newblock In {\em Proceedings of the IEEE/CVF Conference on Computer Vision and Pattern Recognition}, pages 15302--15312, 2021.

\bibitem{caron2021emerging}
Mathilde Caron, Hugo Touvron, Ishan Misra, Herv{\'e} J{\'e}gou, Julien Mairal, Piotr Bojanowski, and Armand Joulin.
\newblock Emerging properties in self-supervised vision transformers.
\newblock In {\em Proceedings of the IEEE/CVF international conference on computer vision}, pages 9650--9660, 2021.

\bibitem{spijkervet2021contrastive}
Janne Spijkervet and John~Ashley Burgoyne.
\newblock Contrastive learning of musical representations.
\newblock {\em arXiv preprint arXiv:2103.09410}, 2021.

\bibitem{poppelbaum2022contrastive}
Johannes P{\"o}ppelbaum, Gavneet~Singh Chadha, and Andreas Schwung.
\newblock Contrastive learning based self-supervised time-series analysis.
\newblock {\em Applied Soft Computing}, 117:108397, 2022.

\bibitem{chen2020simple}
Ting Chen, Simon Kornblith, Mohammad Norouzi, and Geoffrey Hinton.
\newblock A simple framework for contrastive learning of visual representations.
\newblock In {\em International conference on machine learning}, pages 1597--1607. PMLR, 2020.

\bibitem{chen2020improved}
Xinlei Chen, Haoqi Fan, Ross Girshick, and Kaiming He.
\newblock Improved baselines with momentum contrastive learning.
\newblock {\em arXiv preprint arXiv:2003.04297}, 2020.

\bibitem{caron2020unsupervised}
Mathilde Caron, Ishan Misra, Julien Mairal, Priya Goyal, Piotr Bojanowski, and Armand Joulin.
\newblock Unsupervised learning of visual features by contrasting cluster assignments.
\newblock {\em Advances in neural information processing systems}, 33:9912--9924, 2020.

\bibitem{hinton2015distilling}
Geoffrey Hinton, Oriol Vinyals, and Jeff Dean.
\newblock Distilling the knowledge in a neural network.
\newblock {\em arXiv preprint arXiv:1503.02531}, 2015.

\bibitem{zhou2021ibot}
Jinghao Zhou, Chen Wei, Huiyu Wang, Wei Shen, Cihang Xie, Alan Yuille, and Tao Kong.
\newblock ibot: Image {BERT} pre-training with online tokenizer.
\newblock {\em arXiv preprint arXiv:2111.07832}, 2021.

\bibitem{oquab2023dinov2}
Maxime Oquab, Timoth{\'e}e Darcet, Th{\'e}o Moutakanni, Huy Vo, Marc Szafraniec, Vasil Khalidov, Pierre Fernandez, Daniel Haziza, Francisco Massa, Alaaeldin El-Nouby, et~al.
\newblock Dinov2: Learning robust visual features without supervision.
\newblock {\em arXiv preprint arXiv:2304.07193}, 2023.

\bibitem{lee2023evoaug}
Nicholas~Keone Lee, Ziqi Tang, Shushan Toneyan, and Peter~K Koo.
\newblock Evoaug: improving generalization and interpretability of genomic deep neural networks with evolution-inspired data augmentations.
\newblock {\em Genome Biology}, 24(1):105, 2023.

\bibitem{grevsova2023genomic}
Katar{\'\i}na Gre{\v{s}}ov{\'a}, Vlastimil Martinek, David {\v{C}}ech{\'a}k, Petr {\v{S}}ime{\v{c}}ek, and Panagiotis Alexiou.
\newblock Genomic benchmarks: a collection of datasets for genomic sequence classification.
\newblock {\em BMC Genomic Data}, 24(1):25, 2023.

\bibitem{zhang2023dnagpt}
Daoan Zhang, Weitong Zhang, Bing He, Jianguo Zhang, Chenchen Qin, and Jianhua Yao.
\newblock Dnagpt: A generalized pretrained tool for multiple dna sequence analysis tasks.
\newblock {\em bioRxiv}, pages 2023--07, 2023.

\bibitem{dao2022flashattention}
Tri Dao, Daniel~Y. Fu, Stefano Ermon, Atri Rudra, and Christopher R{\'e}.
\newblock Flash{A}ttention: Fast and memory-efficient exact attention with {IO}-awareness.
\newblock In {\em Advances in Neural Information Processing Systems}, 2022.

\bibitem{poli2023hyena}
Michael Poli, Stefano Massaroli, Eric Nguyen, Daniel~Y Fu, Tri Dao, Stephen Baccus, Yoshua Bengio, Stefano Ermon, and Christopher R{\'e}.
\newblock Hyena hierarchy: Towards larger convolutional language models.
\newblock {\em arXiv preprint arXiv:2302.10866}, 2023.

\end{thebibliography}

\clearpage
\appendix

\section{Appendix}

\subsection{GenomicBenchmarks}
GenomicBenchmarks is a recently introduced public benchmark comprising eight distinct regulatory element classification tasks. Seven of these tasks involve binary classification, while the Human Regulatory task specifically entails ternary classification. The sequence lengths within these datasets vary from 2 to 4,776. The median length in each task ranges from 200 to 2,381 (see Table \ref{tab:genomicdata}).

\setlength{\tabcolsep}{2pt}
\begin{table}[h]
\begin{center}
\begin{tabular}{lccc}
\hline\hline\\[-3.5mm]
Dataset & \begin{tabular}{@{}c@{}}Median \\ seq.~len. \end{tabular} & \begin{tabular}{@{}c@{}}training \\ samples\end{tabular} &  \begin{tabular}{@{}c@{}}test \\ samples\end{tabular}\\
\hline\\[-3mm]
Mouse Enhancers & 2381  & 1210   & 242\\ [1mm]

Coding vs Intergenomic &  200 & 75000 & 25000  \\ [1mm]
 
Human vs Worm & 200 & 75000 & 25000 \\ [1mm]

Human Enhancers Cohn & 500 & 20843 & 6948  \\ [1mm]
Human Enhancers Ensembl & 269 & 123872 & 30970\\ [1mm]
Human Regulatory & 401 & 231348 & 57713\\ [1mm]
Human Nontata Promoters & 251 & 27097 & 9034\\ [1mm]
Human OCR Ensembl & 315 & 139804  & 34952\\ [1mm]
\hline\\[-3mm]
\end{tabular}
\end{center}
\caption{Statistics of the GenomicBechmarks datasets.}
\label{tab:genomicdata}
\vspace{-2mm}
\end{table}

\subsection{Computing Overhead Study}
Self-distillation combines masked learning and contrastive learning. Here we investigate the computational overhead brought by contrastive learning for the MTcDNA sequences (1024bp). We measured the time and memory consumed by FinDNA and CM-MNM (i.e., FinDNA with contrastive learning removed). The results, presented in Table \ref{tab:scales}, show that contrastive learning introduces only a small computational overhead.

\begin{table}[h]
\begin{center}
\begin{tabular}{lccc}
\hline\hline\\[-3.5mm]
 & FinDNA & CM-MNM\\
\hline\\[-3mm]
Time (ms) & 95.97 & 95.42\\ [1mm]
Memory(GB) & 5.65 & 4.52 \\ [1mm]
\hline\\[-3mm]
\end{tabular}
\end{center}
\caption{Time and memory FinDNA and CM-MNM consumed for each training batch on the task MTcDNA.}
\label{tab:scales}
\end{table}

\subsection{Student vs Teacher}
 We examined the impact of using teacher and student networks for downstream inference. Both networks were tested on GenomicBenchmarks, and the results are presented in Table \ref{tab:svst}. The table illustrates that the difference between using the teacher and student models is marginal. Despite this, we opted for the teacher network for downstream inference as it exhibited slightly better performance compared to the student network.

\begin{table}[h]
\begin{center}
\begin{tabular}{lcccc}
\hline\hline\\[-3.5mm]
Dataset     &   Student  &  Teacher  \\
\hline\\[-3mm]
Mouse Enhancers & 74.38  & \textbf{76.03} \\ [1mm]

Coding vs Intergenomic &  81.62 & \textbf{81.64} \\ [1mm]
 
Human vs Worm & \textbf{69.58} & 69.24 \\ [1mm]
Human Nontata Promoters & 65.24 & \textbf{65.34}\\ [1mm]
Human OCR Ensembl & 78.79 & \textbf{78.92}  \\ [1mm]
\hline
\end{tabular}
\end{center}
\caption{Performance comparison: using student or teacher networks in the downstream GenomicBenchmarks tasks.}
\label{tab:svst}
\vspace{2mm}
\end{table}

\end{document}